\documentclass[10pt,twocolumn,letterpaper]{article}

\usepackage{cvpr}
\usepackage{times}
\usepackage{epsfig}
\usepackage{graphicx}
\usepackage{amsmath}
\usepackage{amssymb}
\usepackage{amssymb}
\usepackage{caption}
\usepackage{subcaption}
\usepackage{booktabs}
\usepackage{multirow}
\usepackage{float}

\usepackage[breaklinks=true,bookmarks=false]{hyperref}

\cvprfinalcopy 

\ifcvprfinal\fi
\begin{document}

\title{M3DSSD: Monocular 3D Single Stage Object Detector}

\author{Shujie Luo$^{1}$ \space \space Hang Dai$^{3*}$ \space \space Ling Shao$^{3,4}$ \space \space Yong Ding$^{2*}$ \\
%
$^{1}$College of Information Science and Electronic Engineering, Zhejiang University \\ 
$^{2}$School of Micro-Nano Electronics, Zhejiang University \\ 
$^{3}$Mohamed bin Zayed University of Artificial Intelligence, Abu Dhabi, UAE  \\
$^{4}$Inception Institute of Artificial Intelligence, Abu Dhabi, UAE \\
*{\tt\small Corresponding authors\{hang.dai@mbzuai.ac.ae, dingy@vlsi.zju.edu.cn\}. }}

\maketitle
\thispagestyle{empty}

\begin{abstract}
In this paper, we propose a Monocular 3D Single Stage object Detector (M3DSSD) with feature alignment and asymmetric non-local attention. Current anchor-based monocular 3D object detection methods suffer from feature mismatching. To overcome this, we propose a two-step feature alignment approach. In the first step, the shape alignment is performed to enable the receptive field of the feature map to focus on the pre-defined anchors with high confidence scores. In the second step, the center alignment is used to align the features at 2D/3D centers. Further, it is often difficult to learn global information and capture long-range relationships, which are important for the depth prediction of objects. Therefore, we propose a novel asymmetric non-local attention block with multi-scale sampling to extract depth-wise features. The proposed M3DSSD achieves significantly better performance than the monocular 3D object detection methods on the KITTI dataset, in both 3D object detection and bird’s eye view tasks. 
The code is released at \url{https://github.com/mumianyuxin/M3DSSD}.
\end{abstract}

\section{Introduction}

Three-dimensional (3D) object detection enables a machine to sense its surrounding environment by detecting the location and category of objects around it. Therefore, 3D object detection plays a crucial role in systems that interact with the real world, such as autonomous vehicles and robots. The goal of 3D object detection is to generate 3D Bounding Boxes (BBoxes) parameterized by size, location, and orientation to locate the detected objects. 
Most existing methods rely heavily on LiDAR \cite{qi2017pointnet++, shi2019pv, shi2020point, shi2020points, shi2019pointrcnn}, because LiDAR can generate point cloud data with high-precision depth information, which enhances the accuracy of 3D object detection. However, the high cost and short service life make it difficult for LiDAR to be widely used in practice. 
Although binocular camera-based methods \cite{li2019stereo, qin2019triangulation, konigshof2019realtime, chen2020dsgn, chen20173d} achieve good detection results, this is still not a cheap option, and there are often difficulties in calibrating binocular cameras. 
In contrast, the monocular camera is cost-effective, very easy to assemble, and can provide a wealth of visual information for 3D object detection. Monocular 3D object detection has vast potential for applications, such as self-driving vehicles and delivery robots. 

Monocular 3D object detection is an extremely challenging task without the depth provided during the imaging process. To address this, researchers have made various attempts on the depth estimation from monocular images. For instance, \cite{chabot2017deep, barabanau2019monocular} utilize CAD models to assist in estimating the depth of the vehicle. Similarly, a pre-trained depth estimation model is adopted to estimate the depth information of the scene in \cite{vianney2019refinedmpl, bao2019monofenet, wang2019pseudo}. However, such methods directly or indirectly used 3D depth ground-truth data in monocular 3D object detection. Meanwhile, the methods \cite{brazil2019m3d, chen2020monopair} without depth estimation can also achieve high accuracy in the 3D object detection task. In this paper, we propose a 3D object detector for monocular images that achieves state-of-the-art performance on KITTI benchmark \cite{Geiger2012CVPR}.

Humans can perceive how close the objects in a monocular image are from the camera. Why is that? When the human brain interprets the depth of an object, it compares the object with all other objects and the surrounding environment to obtain the difference in visual effect caused by the relative position relationship. For objects of the same size, the bigger, the closer from a fixed perspective. Inspired by this, we propose a novel Asymmetric Non-local Attention Block (ANAB) to compute the response at a position as a weighted sum of the features at all positions. Inspired by \cite{chen20182, zhu2019asymmetric}, we use both the local features in multiple scales and the features that can represent the global information to learn the depth-wise features. The multi-scale features can reduce computational costs. The attentive maps in multiple scales shows an explicit correlation between the sampling spatial resolution and the depth of the objects. 

In one-stage monocular 3D object detection methods, 2D and 3D BBoxes are detected simultaneously. However, for anchor-based methods, there exists feature mismatching in the prediction of 2D and 3D BBoxes. This occurs for two reasons: (1) the receptive field of the feature does not match the shape of the anchor in terms of aspect ratio and size; (2) the center of the anchor, generally considered as the center of the receptive field for the feature map, does not overlap with the center of the object. The misalignment affects the performance of 3D object detection. Thus, we propose a two-step feature alignment method, aiming at aligning the features in 2D and 3D BBox regression. In the first step, we obtain the target region according to the classification confidence scores for the pre-defined anchors. This allows the receptive field of the feature map to focus on the pre-defined anchor regions with high confidence scores. In the second step, we use the prediction results of the 2D/3D center to compute the feature offset that can mitigate the gap between the predictions and its corresponding feature map.

We summarize our contributions as follows:
\begin{itemize}
\setlength\itemsep{0.1mm}
 \item We propose a simple but very efficient monocular 3D single-stage object detection (M3DSSD) method. The M3DSSD achieves significantly better performance than the monocular 3D object detection methods on the KITTI dataset for car, pedestrian, and cyclist object class using one single model, in both 3D object detection and bird’s eye view tasks.
 \item We propose a novel asymmetric non-local attention block with multi-scale sampling for the depth-wise feature extraction, thereby improving the accuracy of the object depth estimation.  
 \item We propose a two-step feature alignment module to overcome the mismatching in the size of the receptive field and the size of the anchor, and the misalignment in the object center and the anchor center. 
\end{itemize}

\section{Related Work}
\label{gen_inst}
In order to estimate depth information in monocular images, researchers have proposed many different approaches. For instance, \cite{xu2018pointfusion, chen2017multi, liang2019multi} utilize point cloud data to obtain accurate 3D spatial information. Pointfusion \cite{xu2018pointfusion} uses two networks to process images and raw point cloud data respectively, and then fuses them at the feature level. MV3D \cite{chen2017multi} encodes the sparse point cloud with a multi-view representation and performs region-based feature fusion. Liang et al. \cite{liang2019multi} exploit the point-wise feature fusion mechanism between the feature maps of LiDAR and images. LiDAR point cloud and image fusion methods have achieved promising performance. However, LiDAR cannot be widely used in practice at present due to its expensive price.

CAD models of vehicles are also used in monocular 3D object detection. Barabanau et al. \cite{barabanau2019monocular} detects 3D objects via geometric reasoning on key points. Specifically, the dimensions, rotation, and key points of a car are predicted by a convolutional neural network. Then, according to the key points’ coordinates on the image plane and the corresponding 3D coordinates on the CAD model, simple geometric reasoning is performed to obtain the depth and 3D locations of the car. Deep MANTA \cite{chabot2017deep} predicts the similarity between a vehicle and a predefined 3D template, as well as the coordinates and visibility of key points, using a convolutional neural network. Finally, given the 2D coordinates of an object's key points and the corresponding 3D coordinates on the 3D template, the vehicle's location and rotation can be solved by a standard 2D/3D matching \cite{lepetit2009epnp}. However, it is difficult to collect CAD models in all kinds of vehicles.

Monocular depth estimation networks are adopted in \cite{vianney2019refinedmpl, ding2019learning, ma2019accurate, bao2019monofenet, xu2018multi, cai2020monocular} to estimate depth or disparity maps. Most of the methods transform the estimated depth map into a point cloud representation and then utilize the approaches based on LiDAR to regress the 3D BBoxes. The performance of these methods relies heavily on the accuracy of the depth map. D4LCN \cite{ding2019learning} proposed a new type of convolution, termed depth-guided convolution, in which the weights and receptive fields of convolution can be automatically learned from the estimated depth.
The projection of the predicted 3D BBox should be consistent with the predicted 2D BBox. This is utilized to build geometric constraints in \cite{naiden2019shift, gahlert2018mb, mousavian20173d} to determine the depth. 
Thanks to the promising performance of convolutional neural networks in 2D object detection, more and more approaches \cite{brazil2019m3d, li2019gs3d, roddick2018orthographic, qin2019triangulation, qin2019monogrnet, liu2020smoke, chen2020monopair, jorgensen2019monocular} have been proposed to directly predict 3D BBoxes using well-designed convolutional neural network for monocular 3D object detection. GS3D \cite{li2019gs3d} proposed a two-stage 3D object detection framework, in which the surface feature extraction is utilized to eliminate the problem of representation ambiguity brought by using a 2D bounding box. M3D-RPN \cite{brazil2019m3d} proposed an anchor-based single-stage 3D object detector that generates both 2D and 3D BBoxes simultaneously. M3D-RPN achieves good performance, but it does not solve the problem of feature misalignment.

\begin{figure*}[t!]
  \centering
  \includegraphics[width=1\textwidth]{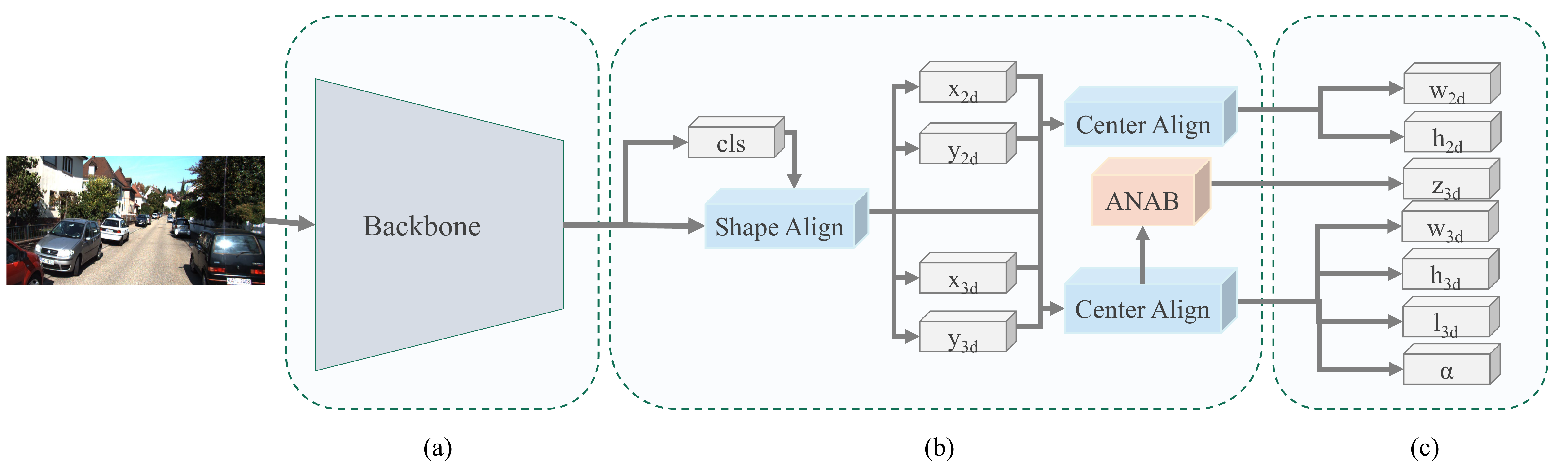}
  \caption{ The architecture of M3DSSD. (a) The backbone of the framework, which is modified from DLA-102 \cite{yu2018deep}. (b) The two-step feature alignment, classification head, 2D/3D center regression heads, and ANAB especially designed for predicting the depth $z_{3d}$. (c) Other regression heads.
} 
  \label{figure1}
  \vspace{-4mm}
\end{figure*}  


\section{Method}
\label{headings}
In this section, we describe the proposed M3DSSD, which consists of four main components: the backbone, the feature alignment, the asymmetric non-local attention block, and the 2D-3D prediction heads, as shown in Fig.~\ref{figure1}. The details of each component are described below.

\subsection{Backbone}
Following \cite{yu2018deep}, we adopt the Deep Layer Aggregation network DLA-102 as the backbone. To adaptively change the receptive field and enhance the feature learning \cite{zhou2019objects, liu2020smoke}, all the convolution in hierarchical aggregation connections are replaced with Deformable Convolution (DCN) \cite{zhu2019deformable}. The down-sampling ratio is set to 8, and the size of the output feature map is $256 \times H/ 8\times W/8$, where $H$ and $W$ are the height and width of the input image.

\subsection{Feature Alignment}
Anchor-based methods often suffer from feature mismatching. On one hand, this occurs if the receptive field of the feature does not match the shape of the anchor in terms of aspect ratio and size. On the other hand, the center of the anchor, generally considered as the center of the receptive field of the feature, might not overlap with the center of the object. The proposed feature alignment consists of shape alignment and center alignment: (1) shape alignment aims at forcing the receptive field of the feature map to focus on the anchor with the highest classification confidence score; (2) center alignment is performed to reduce the gap between the feature on the center of the object and the feature that represents the center of the anchor. Different from previous feature alignment methods \cite{chen20182, wang2019region} that are applied to one-stage object detection via a two-shot regression, the proposed feature alignment can be applied in one shot, which is more efficient and self-adaptive. 

\begin{figure}
  \centering
  \begin{subfigure}[b]{\columnwidth}
    \label{fig.shape1}
    \includegraphics[width=\textwidth]{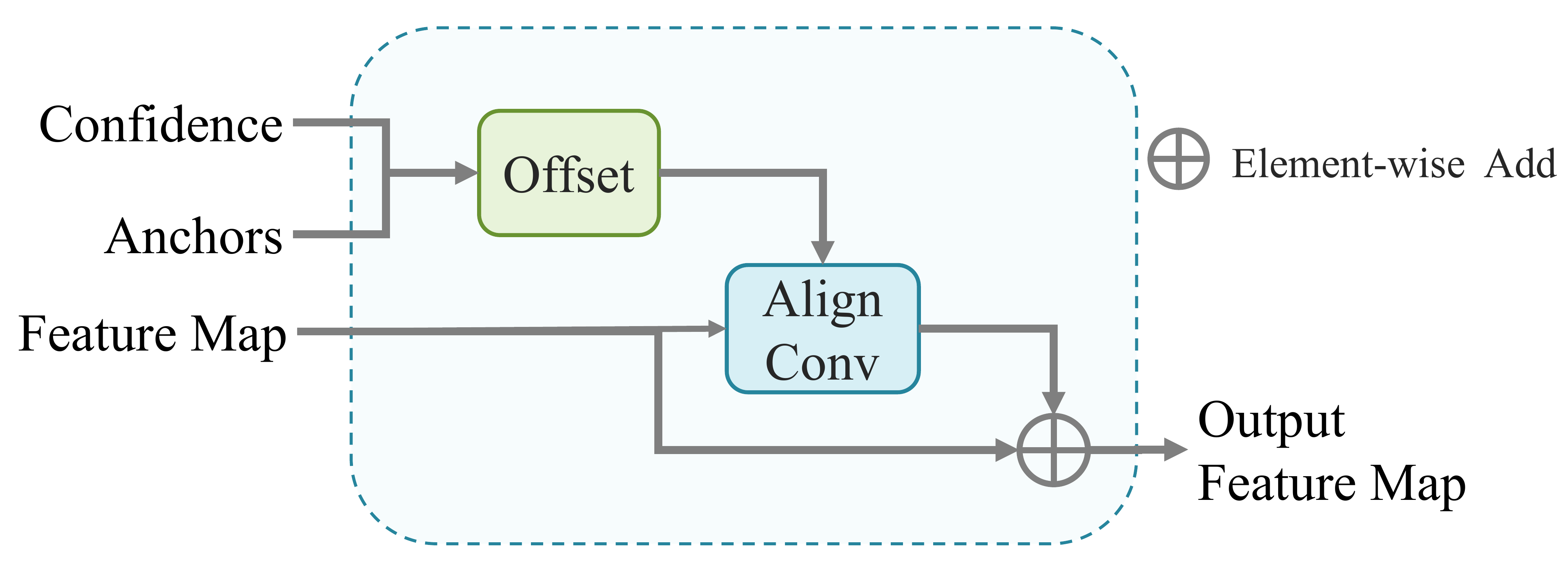}
  \end{subfigure}
  \hfill
  \begin{subfigure}{0.73\columnwidth}
    \label{fig.shape2}
    \includegraphics[width=\textwidth]{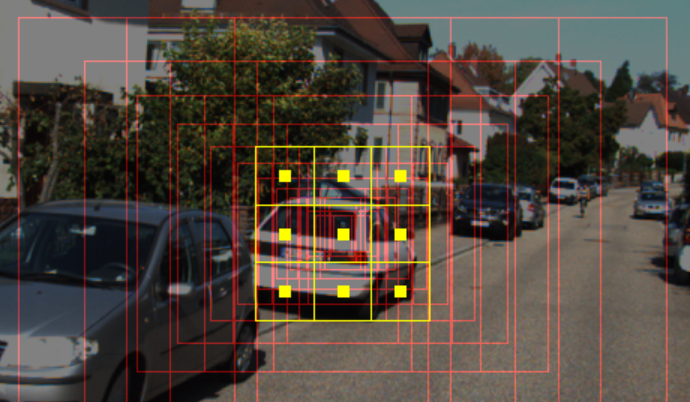}
  \end{subfigure}

  \caption{\small The architecture of shape alignment and the outcome of shape alignment on objects. The yellow squares indicate the sampling location of the AlignConv, and the anchors are in red.}
  \vspace{-4mm}
  \label{fig.shape_align}
\end{figure}

\begin{figure}
  \centering
  \begin{subfigure}[b]{\columnwidth}
    \label{fig.center1}
    \includegraphics[width=\textwidth]{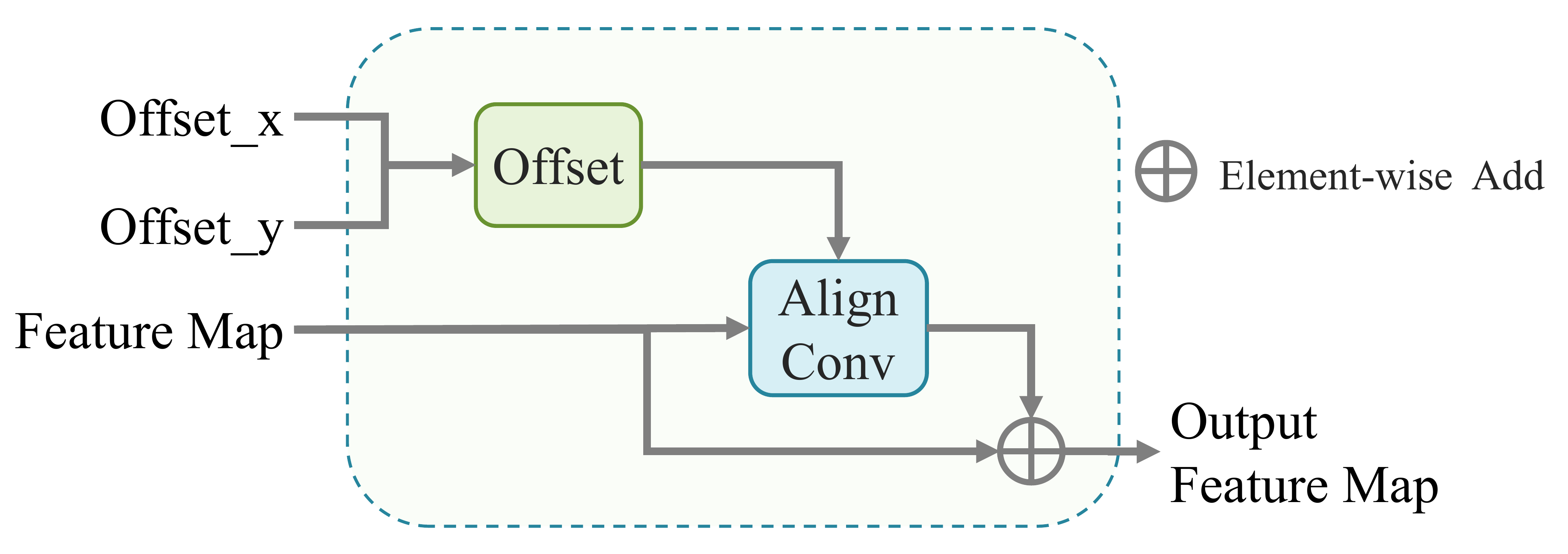}
  \end{subfigure}
  \hfill
  \begin{subfigure}{0.73\columnwidth}
    \label{fig.center2}
    \includegraphics[width=\textwidth]{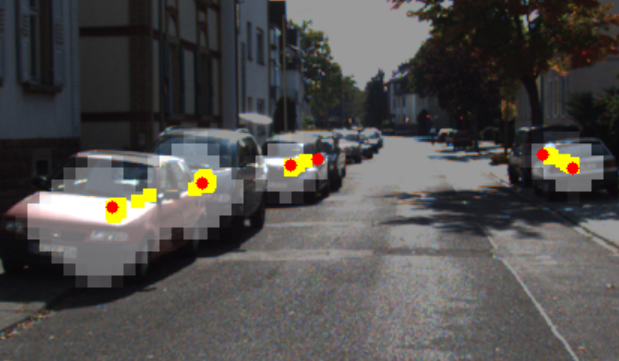}
  \end{subfigure}
  \caption{\small The architectures of center alignment and the outcome of the center alignment. When applying center alignment to objects, the sampling locations on the foreground regions (in white) all concentrate on the centers of objects (in yellow) after center alignment, which are near to the true centers of objects (in red).}
  \vspace{-4mm}
  \label{fig.center_align}
\end{figure}

\textbf{Shape alignment} We can first obtain the foreground region according to the classification results. Then, the receptive fields of the features in the foreground regions can focus on the anchor with the highest confidence scores, as shown in Fig.~\ref{fig.shape_align}. This makes sense because among all the anchors located at the same position, the one with the highest confidence is more likely to remain after the NMS algorithm. We use a convolution termed AlignConv in the implementation of shape alignment and center alignment. AlignConv is similar to the deformable convolution \cite{zhu2019deformable}. The difference is that the offset of the former is computed from the prediction results. The normal convolution can be considered as a special case of AlginConv where the offset equals zero. Unlike the RoI convolution proposed in \cite{chen2019revisiting}, we align the shape of the receptive field or the location of the center in one shot. When performing shape alignment on the feature map with stride $S$, the offset $(O^{sa}_i, O^{sa}_j)$ of the convolution with kernel size $k_h \times k_w$ is defined as:
\begin{eqnarray}
&O^{sa}_i = (\frac{h_{a}}{S \times k_h} - 1) \times (i - \frac{k_h}{2} + 0.5), \\
&O^{sa}_j = (\frac{w_{a}}{S \times k_w} - 1) \times (j - \frac{k_w}{2} + 0.5),
\end{eqnarray}
where $h_{a}, w_{a}$ are the height and the width of the anchor with the highest confidence.

\textbf{Center alignment} The purpose of center feature alignment is to align the feature at the center of the object to the feature that represents the center of the anchor. As shown in Fig.~\ref{fig.center_align}, the prediction results from the 2D/3D center regression are used to compute the offset of the convolution on the feature map with stride $S$:
\begin{eqnarray}
&O^{ca}_i = \frac{y_r}{S},
&O^{ca}_j = \frac{x_r}{S}, 
\end{eqnarray}
where $x_{r}$ and $y_{r}$ are the prediction results of the 2D/3D centers in objects, respectively. As shown in Fig.~\ref{fig.center_align}, when center alignment with a $1 \times 1$ convolutional kernel is applied to the feature map, the sampling position is adaptively concentrated on the center of objects.

\subsection{Asymmetric Non-local Attention Block}
\label{sect:asy}

We propose a novel asymmetric non-local attention block to improve the accuracy of the depth $z_{3d}$ prediction by extracting the depth-wise features that can represent the global information and the long-range dependencies. The standard non-local block \cite{wang2018non} is promising in establishing long-range dependencies, but its computational complexity is $O(N^2C)$, where $N = h \times w$, $h$, $w$ and $C$ indicate the spatial height, width, and channel number of the feature map, respectively. This is very computationally expensive and inefficient compared to normal convolutions. Thus, the applications are limited. The Asymmetric Pyramid Non-local Block \cite{zhu2019asymmetric} reduces the computational cost by decreasing the number of feature descriptors using pyramid pooling. However, pyramid pooling on the same feature map may lead to features with low resolution being replaced with high-resolution features. In other words, there exists redundancy in the computational cost regarding the image resolution. As such, we propose an Asymmetric Non-local Attention Block (ANAB), which can extract multi-scale features to enhance the feature learning with a low computational cost.

\begin{figure}[t!]
    \centering
    \begin{subfigure}{\columnwidth}
        \label{fig:anab}
        \includegraphics[width=\textwidth]{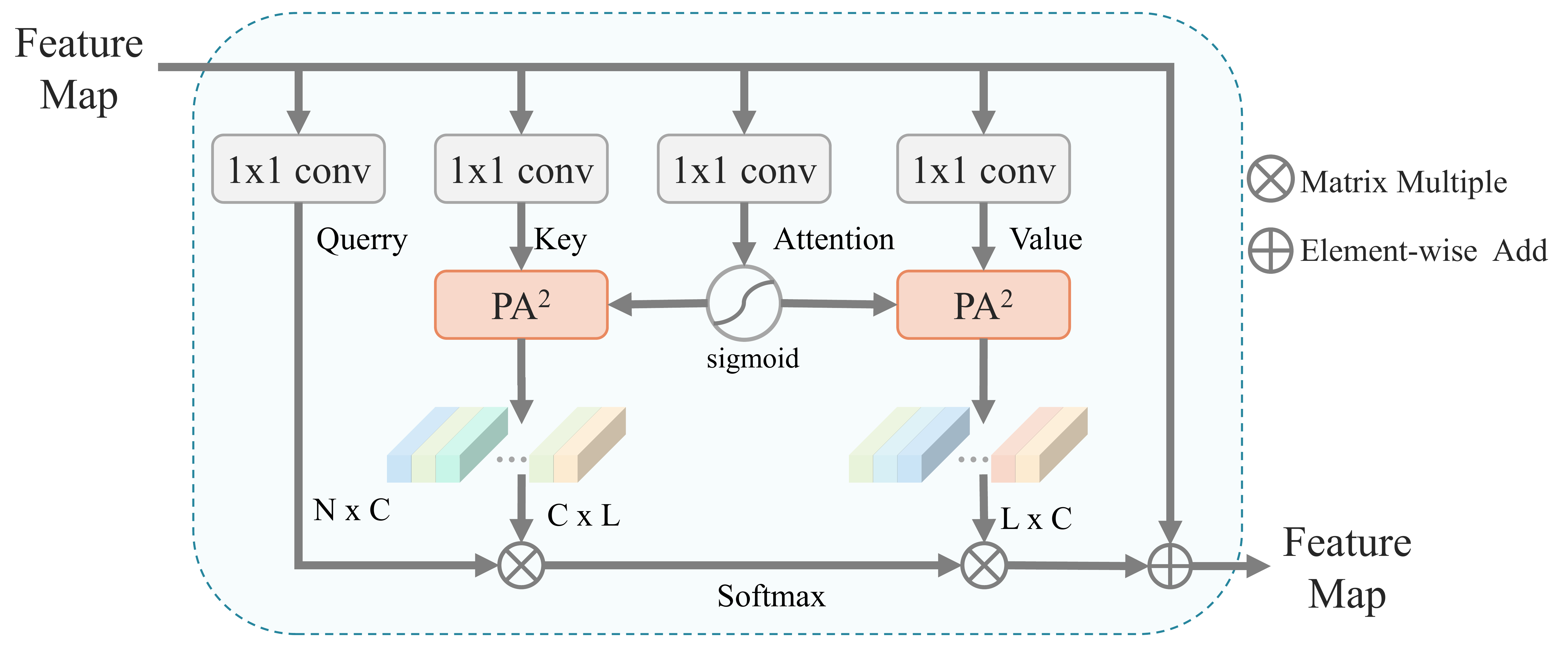}
    \end{subfigure}
    \begin{subfigure}{\columnwidth}
        \label{fig:papa}
        \includegraphics[width=\textwidth]{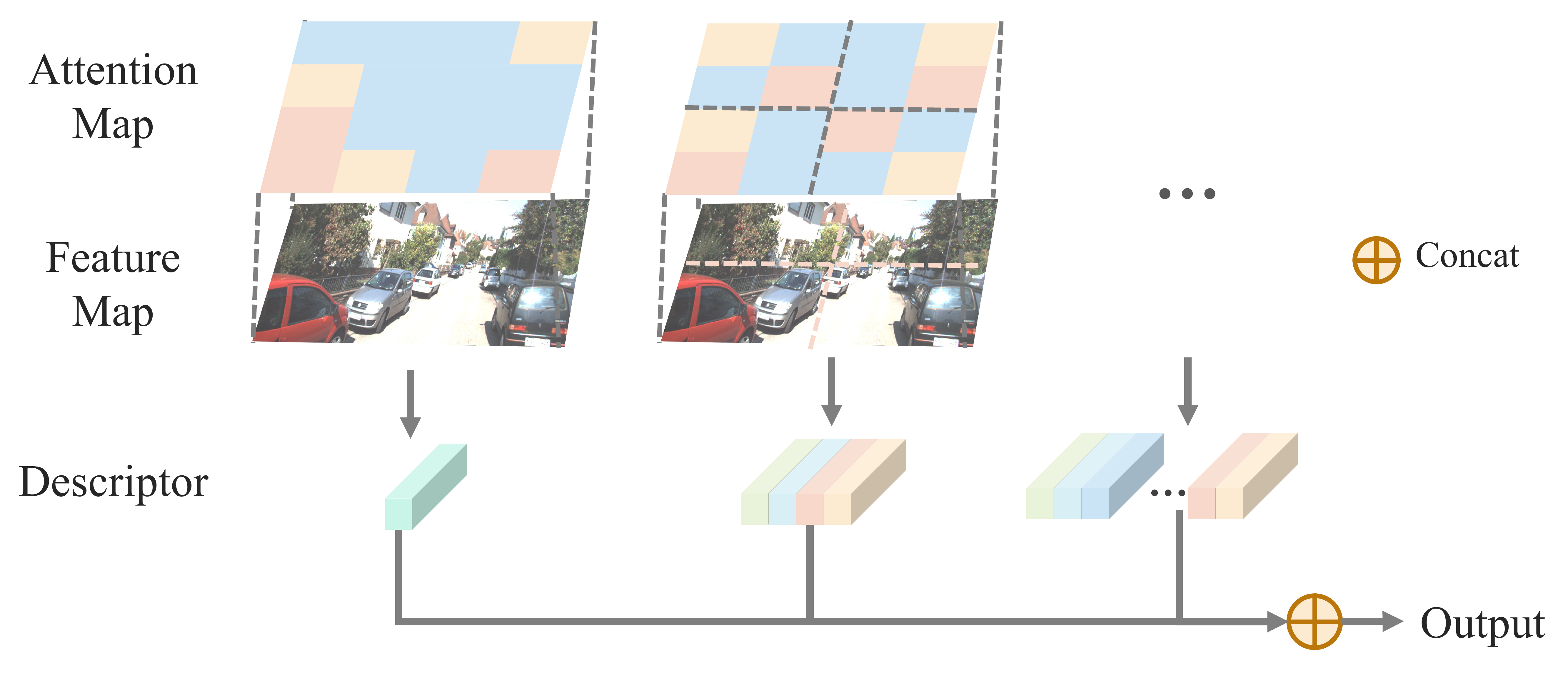}
    \end{subfigure}
    \caption{\small Top: Asymmetric Non-local Attention Block. The key and query branches share the same attention maps, which forces the key and value to focus on the same place. Bottom: Pyramid Average Pooling with Attention ($PA^2$) that generates different level descriptors in various resolutions.}
    \label{fig:pan}
    \vspace{-4mm}
\end{figure}

As shown at the top of Fig.~\ref{fig:pan}, we use the pyramidal features of the $key$ and $value$ branches to reduce the computational cost. The bottom of Fig.~\ref{fig:pan} illustrates the Pyramid Average Pooling with Attention ($PA^2$) module. The different levels of the feature pyramid have different receptive fields, thereby modeling regions with different scales. 
Two matrix multiplications are performed in ANAB. First, the similarity matrix between the reshaped feature matrices $\mathbf{M}_Q$ and $\mathbf{M}_K$ obtained from $querry$ and $key$ is defined as:
\begin{equation}
    \mathbf{M}_{S} = \mathbf{M}_{Q} \times \mathbf{M}_{K}^T, \quad\mathbf{M}_{Q} \in \mathbb{R}^{N\times C}, \mathbf{M}_{K} \in \mathbb{R}^{L\times C}. 
\end{equation}
Then, the softmax function is used to normalize the last dimension of the similarity matrix and multiply it by the reshaped feature matrix $\mathbf M_V$ obtained from $value$ to get the output:
\begin{equation}
    \mathbf{M}_{out} = Softmax(\mathbf{M}_{S}) \times \mathbf{M}_{V} , \quad  \mathbf{M}_{V} \in \mathbb{R}^{L\times C}.
\end{equation}
where $L$ is the number of features after sampling. The standard non-local block \cite{wang2018non} has computational complexity $O(N^2C)$, while the complexity of ANAB is $O(NLC)$. In practice, $L$ is usually significantly smaller than $N$. In our case, we use a four-level downsampling strategy on the feature map $48 \times 160$. The resolution of the four-level feature pyramid is set to $i \in \{1\times1, 4\times4, 8\times8, 16\times16\}$, the sum of which is the total number $L$ of features after downsampling. So $L = 377$ is much smaller than $N = 7680$.

Another effective component of ANAB is the application of the multi-scale attention maps to the $key$ and $value$ branches in $PA^2$ module, as shown at the bottom of Fig. \ref{fig:pan}. 
The motivation is to keep the key information of the origin feature map when greatly reducing the dimensions of matrices $\mathbf{M}_{K}$ and $\mathbf{M}_{V}$ from $N \times C$ to $L \times C$. The spatial attention maps generated by a $1 \times 1$ convolutional layer are used as weights. This module adaptively adjusts the weights to pay more attention to the useful information and suppress the less useful information. 
The attentive map can be treated as a mask performed on multi-scale features. We use the average pooling with attention to downsample the feature maps. Such a weighted average pooling operation offers an efficient way to gather the key features.

\subsection{2D-3D Prediction and Loss}
\textbf{Anchor definition.}
We adopt a one-stage 2D-3D anchor-based network as our detector. To detect the 2D and the 3D BBoxes simultaneously, our predefined anchor contains the parameters of both the 2D BBoxes $[w, h]_{2d}$ and the 3D BBoxes $[z, w, h, l, \alpha]_{3d}$. $\alpha$ is the observation angle of the object that measures the angle at which the camera views the object. 
Compared with the rotation angle of the object, the observation angle is more meaningful for monocular 3D object detection \cite{mousavian20173d}. The dimension of the object is given by  $[w, h, l]_{3d}$. We project the center of the object onto the image plane to encode the 3D location of the object into the anchor:

\begin{equation}
\begin{bmatrix}
X_{p} & Y_{p} & 1
\end{bmatrix} ^\mathrm{T} \cdot Z_{p}
 = \mathbf{K} \cdot 
 \begin{bmatrix}
 X & Y & Z & 1
 \end{bmatrix} ^\mathrm{T}, 
 \label{eqn.1}
\end{equation}
where $(X_p, X_p)$ are the coordinates of the 3D point projected onto the image plane, and $(X, Y, Z)$ are the 3D space coordinates in the camera coordinate system. $K \in \mathbb{R}^{3\times4}$ is the intrinsic camera matrix, which is known at both the training and testing phase.
We obtain the 3D parameters of each anchor by computing the mean of the corresponding 3D parameters of the objects whose intersection over union (IoU) is greater than a given threshold (0.5) with the predefined 2D anchors $[w, h]_{2d}$.

\textbf{Output transformation.}
Given the detection outputs $cls$, $[t_x, t_y, t_w, t_h]_{2d}$ and $[t_x, t_y, t_z, t_w, t_h, t_l, t_\alpha]_{3d}$ for each anchor, the 2D BBox $[X, Y, W, H]_{2d}$ and 3D BBox $[X, Y, Z, W, H, L, A]_{3d}$ can be restored from the output of the detector by:
\begin{align}
&[X, Y]_{2d} = [t_x, t_y]_{2d} \otimes [w, h]_{2d} + [x, y]_{2d} \notag \\
&[W, H]_{2d} = \exp([t_w, t_h]_{2d}) \otimes [w, h]_{2d} \notag \\
&[X_p, Y_p]_{3d} = [t_x, t_y]_{3d} \otimes [w, h]_{2d} + [x, y]_{2d} \notag \\
&[W, H, L]_{3d} = \exp([t_w, t_h, t_l]_{3d}) \otimes [w, h, l]_{3d} \notag \\
&[Z_p, A]_{3d} = [t_z, t_\alpha] + [z, \alpha]_{3d} , \notag \\
\end{align}
where $\otimes$ denotes the element-wise product and $A$ is the rotation angle. During the inference phase, $[X, Y, Z]_{3d}$ can be obtained by projecting $[X_p, Y_p, Z_p]$ back to the camera coordinate system using the inverse operation of Eqn.~\ref{eqn.1}.

\textbf{Loss function.}
We employ a multi-task loss function to supervise the learning of the network, which is composed of three parts: a classification loss, 2D BBox regression loss, and 3D BBox regression loss. The 2D regression and 3D regression loss are regularized with weights $\lambda_1$ and $\lambda_2$:
\begin{equation}
\label{loss}
    L = L_{cls} + \lambda_1 L_{2d} + \lambda_2 L_{3d} ,
\end{equation}
For the classification task, we employ the standard cross entropy loss function:    
\begin{equation}
    L_{cls} = -\log(\frac{\exp(c')}{\sum \exp(c_i)}).
\end{equation}
For the 2D BBox regression task, we use $-\log(IoU)$ as the loss function for the ground-truth 2D BBox $\hat b_{2d}$ and the predicted 2D BBox $b'_{2d}$, similar to \cite{brazil2019m3d}:
\begin{equation}
    L_{2d} = -\log(IoU(b'_{2d}, \hat b_{2d})).
\end{equation}
A smooth L1 loss function is employed to supervise the regression of 3D BBoxes:
\begin{equation}
\begin{array}{l}
    L_{3d} = \sum\limits_{v_{3d} \in P_{3d}} SmoothL_1(v'_{3d}, \hat v_{3d}), \\ P_{3d} = \{t_{x}, t_{y}, t_{z}, t_{w}, t_{h}, t_{l}, t_{\alpha}\}_{3d}.
\end{array}
\end{equation}

\begin{table*}
  \small
  \centering
  \resizebox{1\textwidth}{!}{
  \begin{tabular}{l|c|ccc|ccc}
    \toprule
    \multirow{2}*{Methods}  & \multirow{2}*{Extra}  & \multicolumn{3}{|c|}{$AP_{3d}(val/test) \quad IoU \ge 0.7$} & \multicolumn{3}{c}{$AP_{BEV}(val/test) \quad IoU \ge 0.7$}\\
    {} & {} & Easy & Moderate & Hard & Easy & Moderate & Hard \\
    \midrule
    MonoFENet\cite{bao2019monofenet}  & Depth    &17.54 / 8.35   &11.16 / 5.14   &9.74 / 4.10    &30.21 / 17.03   &20.47 / 11.03  &17.58 / 9.05 \\
    AM3D\cite{ma2019accurate}         & Depth    &32.23 / 16.50  &21.09 / 10.74  &17.26 / 9.52   &43.75 / 25.03   &28.39 / 17.32  &23.87 / 14.91 \\
    D4LCN\cite{ding2019learning}      & Depth    &26.97 / 16.65   &21.71 / 11.72  &18.22 / 9.51   &34.82 / 22.51   &25.83 / 16.02  &23.53 / 12.55 \\
    \midrule
    GS3D\cite{li2019gs3d}             &None      &13.46 / 4.47   &10.97 / 2.90   &10.38 / 2.47   &\qquad- / 8.41  &\qquad- / 6.08  &\qquad- / 4.94 \\
    MonoPSR\cite{ku2019monocular}     &None      &12.75 / 10.76  &11.48 / 7.25   &8.59 / 5.85    &20.63 / 18.33   &18.67 / 12.58   &14.45 / 9.91  \\
    MonoGRNet\cite{qin2019monogrnet}  &None      &13.88 / 9.61   &10.19 / 5.74   &7.62 / 4.25    &\qquad- / 18.19 & \qquad- / 11.17 &\qquad- / 8.73 \\
    SS3D\cite{jorgensen2019monocular} &None      &14.52 / 10.78  &13.15 / 7.68   &11.85 / 6.51   &\qquad- / 16.33 &\qquad- / 11.52 &\qquad- / 9.93 \\
    MonoDIS\cite{simonelli2019disentangling} &None &18.05 / 10.37 &14.98 / 7.94 &13.42 / 6.40 &24.26 / 17.23 &18.43 / 13.19   &16.95 / 11.12 \\
    MonoPair\cite{chen2020monopair}  &None       &\qquad- / 13.04   &\qquad- / 9.99   &\qquad- / 8.65   &\qquad- / 19.28   &\qquad- / 14.83   &\qquad- / \textbf{12.89} \\
    SMOKE\cite{liu2020smoke}         &None       &14.76 / 14.03   &12.85 / 9.76   &11.50 / 7.84   &19.99 / 20.83   &15.61 / 14.49   &15.28 / 12.75 \\
    M3D-RPN\cite{brazil2019m3d}      &None       &20.27 / 14.76   &17.06 / 9.71   &15.21 / 7.42   &25.94 / 21.02   &21.18 / 13.67   &17.90 / 10.23 \\
    RTM3D\cite{li2020rtm3d}          &None       &20.77 / 14.41   &16.86 / 10.34  &16.63 / 8.77   &25.56 / 19.17   &22.12 / 14.20   &20.91 / 11.99 \\
    \midrule
    M3DSSD(ours)  &None    &\textbf{27.77 / 17.51}   &\textbf{21.67 / 11.46}  &\textbf{18.28 / 8.98}   &\textbf{34.51 / 24.15}  &\textbf{26.20 / 15.93}  &\textbf{23.40} / 12.11 \\
    
    \bottomrule
  \end{tabular}
  }
  \caption{AP scores on $val$ and $test$ set of 3D object detection and bird's eye view for cars. }
  \label{tab:ap}
\end{table*}

\section{Experiments}

\begin{figure*}[htbp]
  \centering
  \includegraphics[width=1\textwidth]{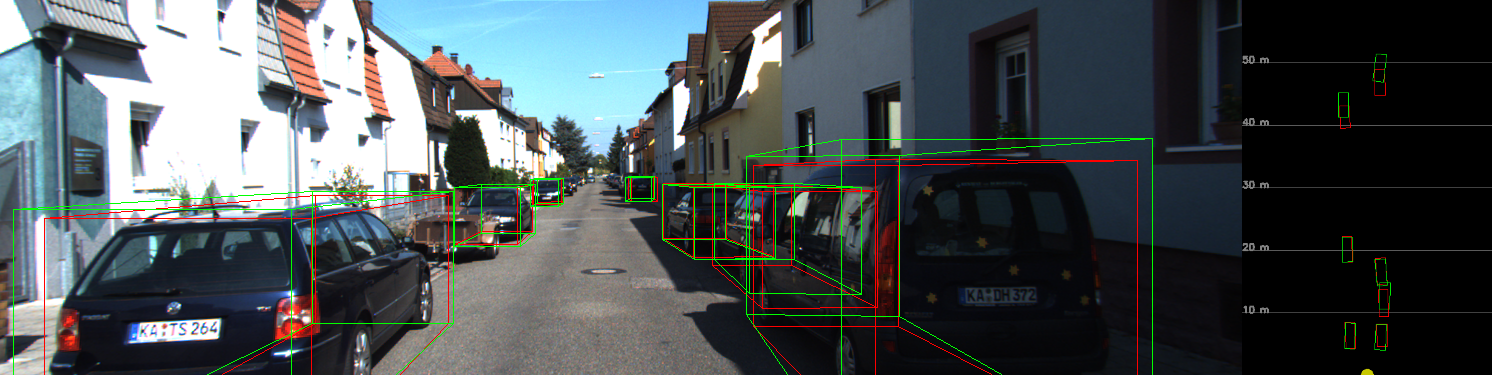}
  \caption{Qualitative results of 3D detection (left) and bird's eye view (right), prediction in green and ground-truth in red.} 
  \label{fig:vis}
  \vspace{-4mm}
\end{figure*}

\subsection{Evaluation Dataset}
We evaluate our framework on the challenging KITTI benchmark for 3D object detection and bird’s eye view tasks. The KITTI dataset contains 7481 images with labels and 7518 images for testing, covering three main categories of objects: cars, pedestrians, and cyclists. We use common split methods \cite{chen20173d} to divide the images with labels into the training set and the validation set. We pad the images to the size of $384 \times 1280$ in both the training and inference phase. In the training phase, in addition to the conventional data augmentation methods of random translation and horizontal mirror flipping, the random scaling operation is applied for monocular images. 

\begin{table*}[htbp]
    \centering
    \begin{tabular}{l|ccc|ccc}
    \toprule
    \multirow{2}*{Methods}  & \multicolumn{3}{|c|}{$Pedestrian \quad AP_{3D}/AP_{bev}$}  & \multicolumn{3}{c}{$Cyclist \quad AP_{3D}/AP_{bev}$}\\
    {}                    &Easy   &Moderate &Hard    &Easy  &Moderate  &Hard\\
    \midrule
    M3D-RPN\cite{brazil2019m3d}           &4.92 / 5.65   &3.48 / 4.05   &2.94 / 3.29   &0.94 / 1.25   & 0.65 / 0.81   & 0.47 / 0.78\\
    D4LCN\cite{ding2019learning}             &4.55 / 5.06   &3.42 / 3.86   &2.83 / 3.59   &2.45 / 2.72   &\textbf{1.67} / 1.82   &1.36 / \textbf{1.79}\\
    SS3D\cite{jorgensen2019monocular}              &2.31 / 2.48   &1.78 / 2.09   &1.48 / 1.61   &\textbf{2.80} / \textbf{3.45}   &1.45 / 1.89   &1.35 / 1.44\\
    \midrule
    M3DSSD(ours)      &\textbf{5.16} / \textbf{6.20}   &\textbf{3.87} / \textbf{4.66}   &\textbf{3.08} / \textbf{3.99}   &2.10 / 2.70    &1.51 / \textbf{2.01}     &\textbf{1.58} / 1.75\\
    \bottomrule
    \end{tabular} 
    \caption{Detection performance for pedestrians and cyclists on $test$ set, at $0.5$ $IoU$ threshold.}
    \label{tab:ped}
    \vspace{-4mm}
\end{table*}

\subsection{Implementation Details}
We implement our model with PyTorch. We adopt the SGD optimizer with momentum to train the network with a CPU E52698 and GPU TITAN V100, in an end-to-end manner, for 70 epochs. The momentum of the SGD optimizer is set to 0.9, and weight decay is set to 0.0005. The mini-batch size is set to 4. The learning rate increases linearly from 0 to the target learning rate of 0.004 in the first epoch and then decreases to $4\times 10^{-8}$ with cosine annealing. Terms $\lambda_1$ and $\lambda_2$ in Eqn.~\ref{loss} are both set to 1.0. 
We lay 36 anchors on each pixel of the feature map, the size of which increases from 24 to 288 following the exponential function of $24 \times 12^{i/11}, i \in \{0, 1, 2, \dots , 11\}$, and the aspect ratio is set to $\{0.5, 1.0, 1.5\}$. We apply online hard-negative mining by sampling the top $20\%$ high loss boxes in each minibatch in the training phase. 
In the inference phase, we apply NMS with 0.4 IoU criteria on the 2D BBox and filter out the objects with a confidence lower than 0.75. The post-optimization algorithm proposed in \cite{brazil2019m3d} is used to make the rotation angle more reasonable. 
The algorithm uses projection consistency to optimize the rotation angle. The rotation angle is optimized iteratively to minimize the L1 loss of the projection of the predicted 3D BBox and the predicted 2D BBox.

\subsection{Performance Evaluation}
We set the network after removing the feature alignment module and ANAB from M3DSSD as the baseline. More specifically, for the baseline, the feature map output from the backbone is directly used for classification and 2D BBox regression and 3D BBox regression.

We evaluate our framework on the KITTI benchmark for both bird’s eye view and 3D object detection tasks. The average precision (AP) of Intersection over Union (IoU) is used as the metric for evaluation in both tasks and it is divided into easy, moderate, and hard according to the height, occlusion, and truncation level of objects. Note that the official KITTI evaluation has been using $AP|_{R40}$ with 40 recall points instead of $AP|_{R11}$ with 11 recall points since October 8, 2019. However, most previous methods evaluated on the validation used $AP|_{R11}$. Thus, we report the $AP|_{R40}$ for the test dataset and $AP|_{R11}$ for the validation dataset for a fair comparison. We set the threshold of IoU to 0.7 for cars and 0.5 for pedestrians and cyclists as the same as the official settings. Fig.~\ref{fig:vis} shows qualitative results for 3D object detection and bird's eye view. The detection results and depth predictions are less accurate with further distance. The videos of 3D object detection results and the additional results can be found in the supplemental material. 
 

\textbf{Bird's eye view.}
The bird’s eye view task is to detect objects projected on the ground, which is closely related to the 3D location of objects. The detection results for cars on both the $val$ and $test$ set are reported in Tab.~\ref{tab:ap}. M3DSSD achieves state-of-the-art performance on the bird’s eye view task compared to approaches with and without depth estimation. Our method has significant improvement compared to the methods without depth estimation. 

\begin{figure}[htb]
  \centering
  \includegraphics[width=0.8\columnwidth]{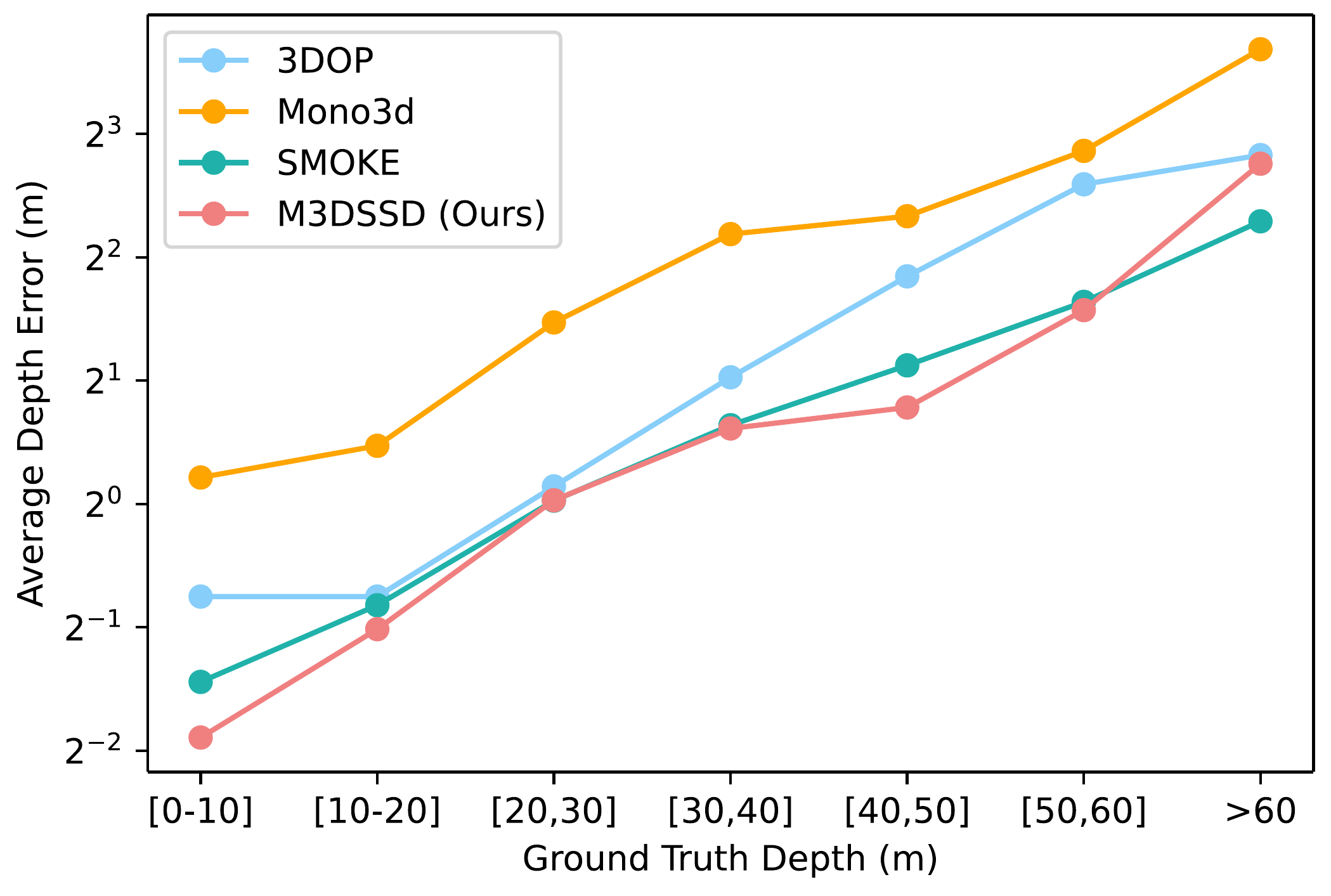}
  \caption{The average depth estimation error varies along with the ground truth depth. Best viewed in color.} 
  \label{figure.depth_err}
  \vspace{-4mm}
\end{figure}

\begin{table}[htbp]
  \centering
  \resizebox{1\columnwidth}{!}{
  \begin{tabular}{l|ccc}
    \toprule
    \multirow{2}*{Methods}  & \multicolumn{3}{|c}{$AP_{3d}/AP_{BEV}\quad IoU \ge 0.7 $}\\
    {}                    &Easy   &Mod. &Hard   \\
    \midrule
     
    Baseline w/ ANAB \dag            & 25.70 / 33.48        &19.02 / 24.79 &17.31 / 20.15  \\ 
    \dag w/ Shape Alignment   & 27.26 / 33.64        &21.56 / 25.24 &18.07 / 22.81  \\
    \dag w/ Center Alignment  & 27.33 / \textbf{34.85} &21.51 / 25.96 &18.03 / 23.26  \\
    \dag w/ Full Alignment    & \textbf{27.77} / 34.51 &\textbf{21.67} / \textbf{26.20} &\textbf{18.28} / \textbf{23.40}  \\
    \bottomrule
  \end{tabular}
  }
  \caption{Ablation study on feature alignment.}
  \label{tab:feature alignment}
\end{table}

\begin{table*}[t]
    \centering
    \resizebox{0.9\textwidth}{!}{
    \begin{tabular}{l|ccc}
        \toprule
        \multirow{2}*{Methods}  & \multicolumn{3}{|c}{$AP_{3d}/AP_{BEV} \quad IoU \ge 0.7 $} \\
        {}                    &Easy        &Moderate         &Hard    \\
        \midrule
         baseline            &23.40 / 28.66    &18.32 / 23.53    &16.62 / 19.54 \\
         ANB                 &23.65 / 29.19    &18.47 / 23.65    &16.54 / 19.50  \\
         ANAB  &\textbf{25.70} / \textbf{33.48}  &\textbf{19.02} / \textbf{24.79}  &\textbf{17.31} / \textbf{20.15} \\
        \bottomrule
    \end{tabular}
    
    \begin{tabular}{||l|c|c}
        \toprule
        \multirow{2}*{Methods} & \multirow{2}*{GPU time}  & \multirow{2}*{GPU memory}\\
        {}   &  (ms)  & (Gbyte)\\
        \midrule
         Non-local \cite{wang2018non}     &   5.89 / 104.12         & 1.97 / 15.67 \\
         ANB                              &   \textbf{1.68} / \textbf{5.92}  & \textbf{1.09} / \textbf{1.43} \\
         ANAB                             &   1.86 / 6.76           & 1.22 / 1.91 \\
        \bottomrule
    \end{tabular}}
    
    \caption{Ablation study on non-local blocks with detection accuracy, GPU time, and memory for different input sizes.}
    \label{tab:anab}
\end{table*}

\begin{figure}[htb]
  \centering
  \includegraphics[width=0.8\columnwidth]{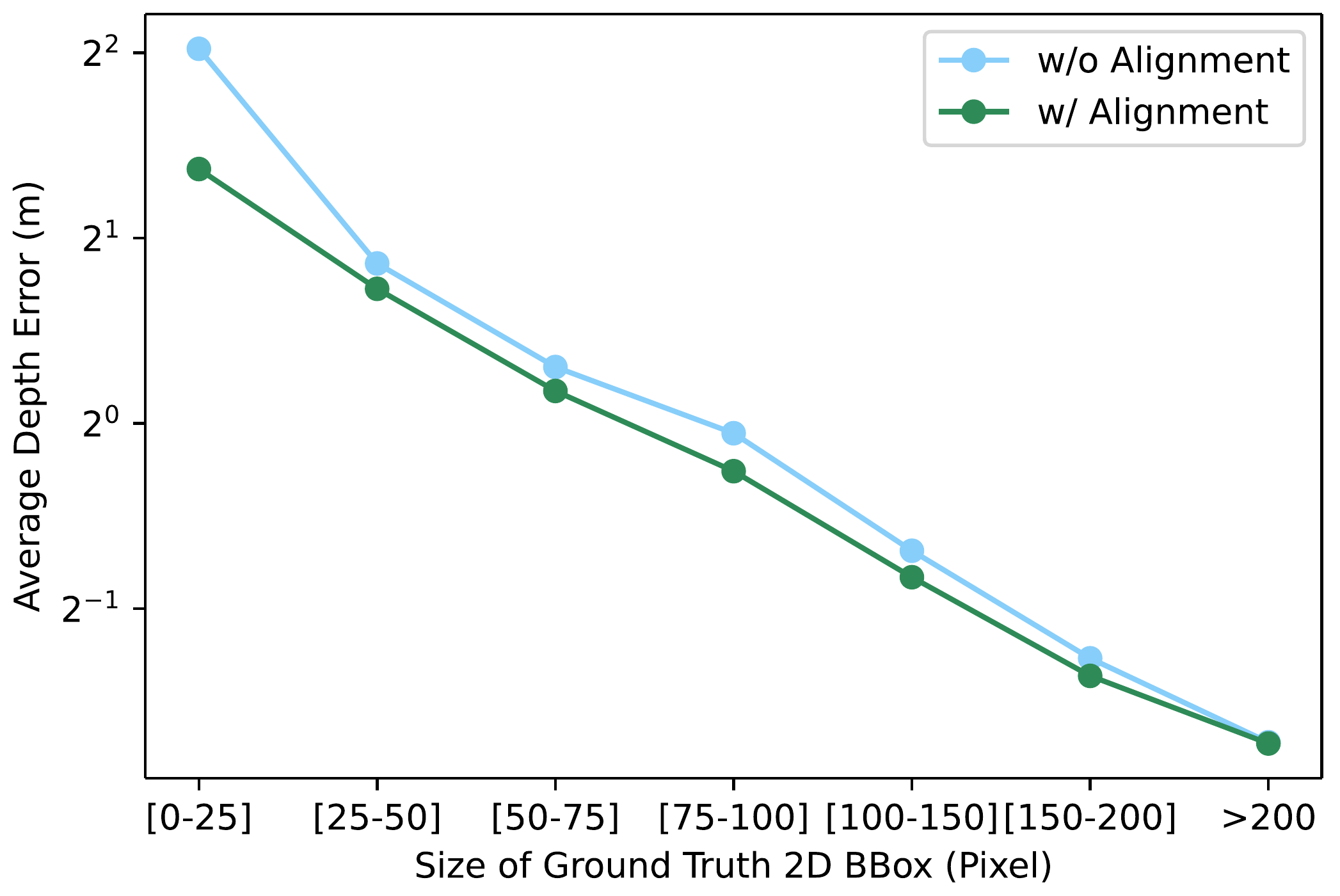}
  \caption{The average depth estimation error varies along with the size of objects that is the average of the length and the width of the 2D BBox. Best viewed in color.} 
  \label{figure.depth_err_alg}
  \vspace{-4mm}
\end{figure}

\begin{figure*}[htb]
  \flushright
  \includegraphics[width=0.93\textwidth]{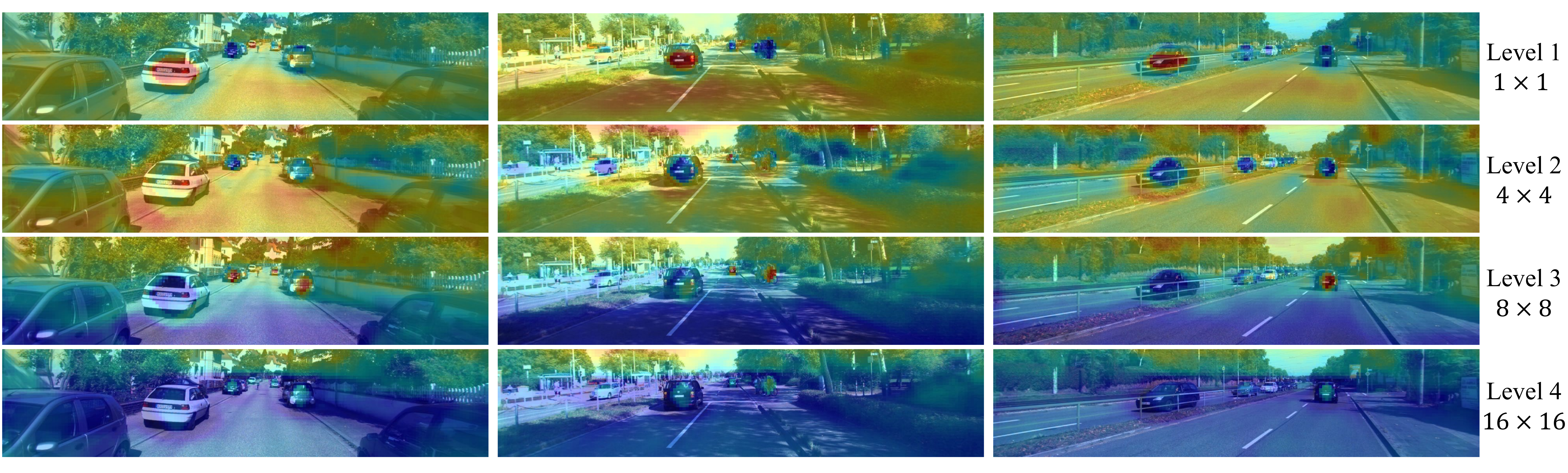}
  \caption{Visualization for attention maps in $PA^2$ with a four-level feature pyramid $\{1\times1, 4\times4, 8\times8, 16\times16\}$.} 
  \label{figure.anab_vis}
  \vspace{-4mm}
\end{figure*}
\textbf{3D object detection for cars.}
The 3D object detection task aims to detect 3D objects in the camera coordinate system, which is more challenging than the bird’s eye view task due to the additional y-axis. Compared with the approaches without depth estimation, Tab.~\ref{tab:ap} shows that M3DSSD achieves better performance in both the $val$ and $test$ set. Note that M3DSSD is better than most of the approaches with depth estimation. Further, our method achieves competitive performance against D4LCN that adopts a pre-trained model for depth estimation \cite{ding2019learning}. 

Fig.~\ref{figure.depth_err} shows the average depth estimation error with the different ground truth depth ranges \cite{liu2020smoke}. We compared our proposed method with SMOKE \cite{liu2020smoke}, Mono3D \cite{chen2016monocular} and 3DOP \cite{chen20173d} on the same validation set. Fig.~\ref{figure.depth_err} demonstrates that the proposed M3DSSD achieves better performance at all distance ranges, except for the distance greater than 60m, where the number of samples is usually small. 

\textbf{3D object detection for pedestrians and cyclists.}
Compared with cars, 3D object detection for pedestrians and cyclists is more challenging. This is because the size of pedestrians and bicycles is relatively small. In addition, people are non-rigid bodies, and their shapes vary a lot, thereby making it difficult to locate pedestrians and cyclists.
We report the detection results for pedestrians and cyclists on the test set of the KITTI benchmark in Tab.~\ref{tab:ped}. Since some methods did not report the pedestrian and the cyclist results, we compare our model with M3D-RPN \cite{brazil2019m3d}, D4LCN \cite{ ding2019learning}, and SS3D \cite{jorgensen2019monocular}. Our model achieves competitive performance in both 3D detection and bird’s eye view tasks for pedestrians and bicycles, especially for the pedestrian category. Note that we train only one single model to detect the three object classes simultaneously. 

\subsection{Ablation Study}
\textbf{Feature alignment.} 
We evaluate the feature alignment strategies, including shape alignment, center alignment, and full alignment (both center alignment and shape alignment). 
As shown in Tab. \ref{tab:feature alignment}, that the proposed shape alignment, center alignment, and full alignment achieve better results compared to the case without alignment. 

Fig.~\ref{figure.depth_err_alg} illustrates the average depth estimation error varies with the size of objects for the model with and without feature alignment. The x-axis is set as the size of the 2D BBox $(w_{2d}+h_{2d})/2$. It shows that the proposed feature alignment module is effective on objects of different sizes, especially for the small objects in $[0 - 25]$. This also explains why M3DSSD outperforms other methods in small object detection such as pedestrians and cyclists.

\textbf{Asymmetric non-local attention block.}
We compare the Asymmetric Non-local Block (ANB), and our proposed Asymmetric Non-local Attention Block (ANAB), which applies pyramid average pooling on the feature map with attentions. We use the same sampling size for both methods. Tab.~\ref{tab:anab} shows that the network with ANAB achieves the best performance. With a similar computational time, the proposed ANAB has better detection accuracy than ANB. Meanwhile, both methods cost much less GPU time and memory than the standard non-local block \cite{wang2018non}. The attention module costs a little more consuming time with significant improvement, especially in easy tasks. Tab. \ref{tab:anab} on the right shows the GPU time and memory regarding the input size $[1, 256, 48, 160]$ and $[1, 256, 96, 320]$. This shows that the computational cost is closer to the theoretical analysis in Sect.~\ref{sect:asy} with a larger input size. ANAB has extra pooling layers, convolutional layers, and an element-wise multiplication, which are not considered in the theoretical analysis.

In ANAB, the attention maps are assigned to the multi-scale pooling operations for the depth-wise feature extraction. Fig.~\ref{figure.anab_vis} shows that the attention map for $1 \times 1$ feature pyramid has larger weights on the objects which are close to the camera, while the attention map for the higher-level feature pyramid assigns larger weights on the objects that are away from the camera. The attention maps in different levels show a correlation between the resolution of the feature pyramid and the object depth. This lies in the fact that the feature pyramid with low resolution has a large receptive field that is sensitive to the object in large size, while the feature pyramid with high resolution has a small receptive field that is sensitive to the object in small size. For the size of the same-class object from a fixed perspective, the smaller, the farther.
The depth-wise attention maps enhance the capability of perceiving the depth of objects, thereby improving the performance of object depth estimation.

\section{Conclusion}

In this work, we propose a simple and very effective monocular single-stage 3D object detector. We present a two-step feature alignment approach to address the feature mismatching, which enhances the feature learning for object detection. The asymmetric non-local attention block enables the network to extract depth-wise features, which improves the performance of the depth prediction in the regression head. Compared to the methods with or without the estimated depth as an extra input, M3DSSD achieves better performance on the challenging KITTI dataset for car, pedestrian, and cyclist object class using one single model, for both bird’s eye view and 3D object detection.

\textbf{Acknowledgement:} This work was supported in part by the National Key Research and Development Program of China (2018YFE0183900). Hang Dai would like to thank the support from MBZUAI startup fund (GR006).

{\small
\bibliographystyle{ieee_fullname}
\bibliography{ms}
}

\end{document}